\documentclass{sig-alternate}
\usepackage{epsfig}
\usepackage{authblk}
\usepackage{graphicx}
\usepackage{caption3}
\usepackage{lipsum}
\usepackage{amssymb}

\usepackage{amsthm}

\usepackage{algorithm}
\usepackage{algorithmic}
\usepackage{comment}
\usepackage{textcomp}

\newtheorem{theorem}{Theorem}[section]
\newtheorem{corollary}[theorem]{Corollary}

\newtheorem{lemma}[theorem]{Lemma}

\newcommand{\QED}{\mbox{}\hfill \rule{3pt}{8pt}\vspace{10pt}\par}

\graphicspath{{./figures/}}

\begin{document}

\title{Fast Approximate Matching of Cell-Phone Videos for Robust Background Subtraction}

\author{\textbf{Raffay Hamid, Atish Das Sarma, Dennis DeCoste, Neel Sundaresan}\\ \vspace{-0.45cm}{{raffay@gmail.com, atish.dassarma@gmail.com, ddecoste@ebay.com, nsundaresan@ebay.com}}
\\ \vspace{-0.125cm}\textbf{eBay Research Laboratory, San Jose, CA. USA.}}


\maketitle

\begin{abstract}

We identify a novel instance of the background subtraction problem that focuses on extracting near-field foreground objects captured using handheld cameras. 
%
%
Given two user-generated videos of a scene, one with and the other without the foreground object(s), our goal is to efficiently generate an output video with only the foreground object(s) present in it. We cast this challenge as a spatio-temporal frame matching problem, and propose an efficient solution for it that exploits the temporal smoothness of the video sequences. We present theoretical analyses for the error bounds of our approach, and validate our findings using a detailed set of simulation experiments. Finally, we present the results of our approach tested on multiple real videos captured using handheld cameras, and compare them to several alternate foreground extraction approaches.

\end{abstract}

\section{Introduction}

\noindent Over the years, there has been a tremendous increase in the number of videos recorded from cell-phones and other handheld mobile devices. A majority of these videos capture a set of near-field objects in visually cluttered environments. Automatic extraction of these foreground objects in such videos is an important problem, with potential impacts on a variety of different application areas.

For videos captured using handheld cameras, it is relatively easy to record a few frames of the scene without the foreground objects present, to coarsely characterize what the background looks like. We argue that this extra, albeit approximate, information about the background scene can be very useful to extract foreground objects in such videos. 

The motivation of our novel problem setting is grounded in several real-world applications. In online commerce for instance, thousands of sellers use mobile phones to capture videos of their products they want to sell online. Most of these sellers are amateur videographers who lack the skills or the equipment to capture good quality videos of their products. An important factor adversely affecting their video quality is the background scene clutter. As improving the quality of product videos increases the likelihood of selling the products, these users are naturally incentivized to provide a short additional video of the background scene in return for having product videos with improved quality. 

Similarly on filming rigs, it is a standard practice to capture background and foreground scene videos along carefully planned camera paths using expensive computerized dollies to generate various special effects. By enabling robust and efficient background-foreground video matching using only hand-held cameras, our work attempts to make this process more efficient and less expensive. Furthermore, our work attempts to open up avenues for amateur videographers who cannot afford costly gear, to generate these special effects using only their handheld mobile cameras.

The precise objective of our work is, given two handheld videos, one with and the other without foreground object(s), to efficiently generate a video with only the foreground object(s) shown. Note that if camera paths for both the foreground and the background videos in our problem were identical, we could simply perform a pixel-wise background subtraction. On the other hand, if the two camera paths were entirely different with no scene overlap, the background video would be redundant and our problem would transform to an object tracking challenge. In practice however, one is somewhere in the middle of this spectrum with some amount of partial path overlap between the two videos. The challenge therefore is, for each foreground frame, to find a set of background frames with sufficient scene overlap, such that with the appropriate frame-alignment, they could be used to subtract out the background pixels from the foreground frame.

A naive approach to this spatio-temporal frame matching problem requires operations quadratic in the number of the foreground and background frames. For only one minute long background and foreground videos captured at $30$ frames per second, a naive approach requires more than $3$ million pair-wise frame comparisons. Even with an optimized implementation exploiting heterogeneous compute architecture, each comparison can take up to $100$ ms, which would result in the entire process taking several days to complete. It is therefore crucial that for each foreground frame, we find the relevant background frames more efficiently.

\vspace{0.175cm}\noindent The \textbf{main contributions} of our work are:

\vspace{0.1cm}\noindent ${\bullet}$ We propose an approximation algorithm for efficient spatio-temporal matching of videos captured using hand-held cameras. Our algorithm permits a significantly improved computational complexity compared to an exhaustive approach.

\vspace{0.1cm}\noindent ${\bullet}$ We present a systematic way to quantify bounds on the error incurred by our approximate matching algorithm, and corroborate our theoretical findings by a set of controlled simulation experiments.

\vspace{0.1cm}\noindent ${\bullet}$ We present a computational framework using our proposed frame matching algorithm to efficiently subtract the scene background given a background and a foreground video.

\vspace{0.1cm}\noindent ${\bullet}$ We validate the robustness of our framework on numerous real videos, and show the improved performance of our technique compared to multiple alternate approaches.

\vspace{0.1cm}\noindent In the following, we start by going over the relevant previous works identifying how our work is different from them. We then concretely formalizing our spatio-temporal frame matching problem. We use the insights from our formulation to make appropriate choices for our computational framework explained in $\S$~\ref{sec:comp_framework}.



\section{Related Work}

\noindent Foreground extraction in videos has previously been explored in great detail~\cite{stauffer1999adaptive}~\cite{toyama1999wallflower},
however most of this work has either focused on static camera(s), or for cameras with limited motion~\cite{hayman2003statistical}~\cite{bevilacqua2006high}. To overcome this challenge, there have been several efforts to look at the problem purely from a tracking based
perspective~\cite{isard2001bramble}~\cite{comaniciu2000real}. While these approaches
can successfully localize object position, they are not geared for accurately extracting object silhouettes.

A solution for this challenge is to use tracking and segmentation in closed-loop form~\cite{sun2006background}~\cite{papoutsakis2010object}. However, such approaches either require human intervention to
provide object silhouette, or assume specific lighting conditions.

Another method to estimate object silhouette to initialize tracking is to segment the scene into multiple motion
layers~\cite{sheikh2009background}~\cite{torr2001integrated}. However, such methods are limited to non-static scenes with
distinct object-level motion boundaries.

There have been some recent attempts to combine appearance and stereo cues for foreground extraction. Most of these methods however perform full sequence calibration~\cite{KowdleSS12}, or assume that the background scene can be approximated using piecewise planar geometry~\cite{GranadosKTKT12}. There have also been attempts to extract foreground object(s) based on frame saliency~\cite{itti1998model}~\cite{harel2006graph} and co-segmentation~\cite{meng2010object}.

Recently there has been work on utilizing the general background appearance
sampled sparsely over space and time from unregistered images~\cite{aghazadeh2012multi}~\cite{aghazadeh2011novelty}. We show
in $\S$~\ref{ss:mathing_not enough} however that this model does not generalize for handheld video cameras as their unconstrained
motion requires quite a dense sampling of background scene appearance, resulting in prohibitively high capture and computational cost. Similar approaches for scene alignment have been explored for videos~\cite{sand2004video}~\cite{carceroni2004linear}~\cite{caspi2002spatio}. However, unlike our work, they assume same or near identical camera paths. 
\section{Problem Setup}
\label{sec:problem}
\begin{figure}[t]
\begin{center}
   \includegraphics[width=0.8\linewidth]{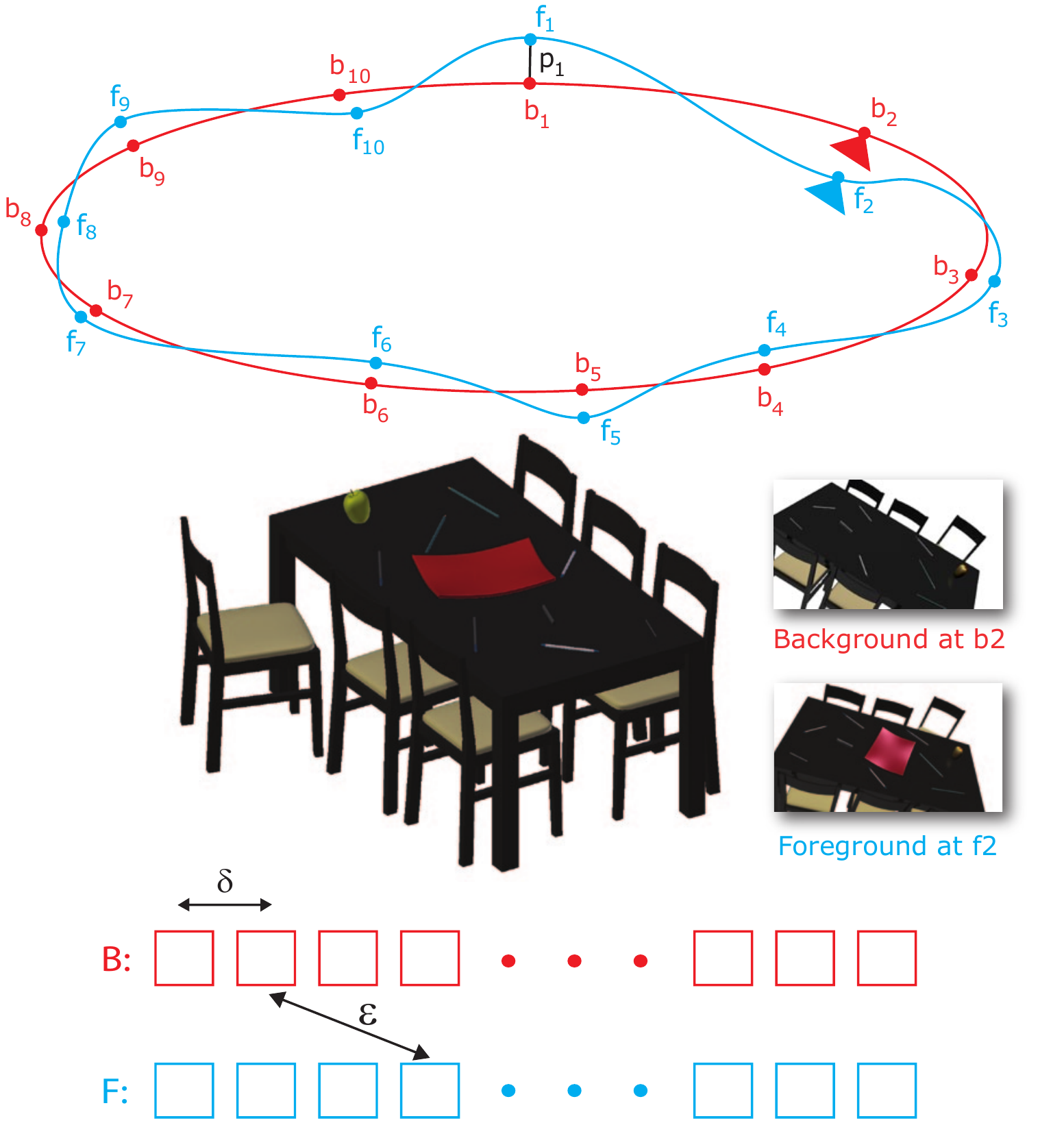}
\end{center}
   \caption{An example $3$-D scene is captured by a handheld camera tracing two different paths, $f$ and $b$, one with and
   the other without a foreground object (red tray). This results in a foreground and a background video, $\bar{f}$
   and $\bar{b}$ respectively. Example frames from these videos for this scene are also shown at $f_{2}$ and $b_{2}$.}
\label{fig:problem_statement_fig}
\end{figure}

\noindent Consider Figure~\ref{fig:problem_statement_fig}, where a $3$-D scene is shown to be captured by a handheld camera
tracing two different paths. For one of these captures, the scene contains a (set of) foreground object(s) resulting in a
foreground video. For the second capture, a background video is recorded without the foreground object(s).
Let $n$ and $m$ denote the number of foreground and background video frames respectively. Let $\bar{f} = \{f_1, f_2, \ldots, f_n\}$ and $\bar{b} = \{b_1, b_2, \ldots, b_m\}$ denote the foreground and background frame sequences respectively. We use $d(x,y)$ to denote the distance between frames $x$ and $y$, the exact notion of which is
formally defined in $\S$~\ref{ss:key_point_detection}.

\subsection{Problem Assumptions}
\label{ss:problem_assumptions}
\noindent We make the following assumptions to setup our problem.

\subsubsection{Path Smoothness}
\label{ssec:path_smoothness}
\noindent We assume that consecutive frames in a video are similar to each other, \textit{i.e.}, the path traced by the camera during a capture is generally \textit{smooth}. Given the high capture rate of modern mobile devices, this assumption is well-justified. Formally, $\forall$ $i$ $(1\leq i<n)$, we have $d(f_i, f_{i+1})\leq \delta$, and $\forall$ $j$ $(1\leq j<m)$, we have $d(b_j, b_{j+1})\leq \delta$, for $\delta > 0$.

\subsubsection{Capture Completeness}
\label{sssec:capture_completeness}
\noindent Given our problem setup, it is likely that several foreground frames would have strong background matches, \textit{i.e.}, for such foreground frames the background capture would be \textit{complete}. For the sake of clarity, we begin with the assumption that capture completeness applies $\forall f \in \bar{f}$. We address the scenario when this is not the case in $\S$~\ref{ssec:foreground_tracking}. Formally, for each $i$ between $1\leq i\leq n$, $\exists$ $j$ between $1\leq j\leq m$ such that $d(f_i,b_j)\leq \epsilon$. Note that here $\epsilon$ is significantly larger than $\delta$ (as the object of interest is present only in the foreground video), yet typically small when the background video is fairly exhaustive.

\subsection{Problem Formulation}
\noindent Given $\bar{f}$ and $\bar{b}$, $\forall$ $1\leq i\leq n$, we want to find a corresponding $1\leq j\leq m$ to
minimize $d(f_i, b_j)$. More formally, we want to identify function $\pi: [1,n]\rightarrow [1,m]$ such that
\texttt{avg}$_{i=1}^{i=n}{d(f_i,b_{\pi(i)})}$ is minimized. Note that here we use the average cost; however one could define
the objective using $\max$ (or equivalent functions) and similar results hold.

\vspace{-0.125cm}\section{Spatio-Temporal Matching}
\label{sec:Spatio-Temporal-Frame-Matching}

%

\noindent In the following, we first present a naive algorithm for frame matching between the foreground and background videos, followed by our proposed approximate algorithm that guarantees a significant speedup while only incurring minimal accuracy cost. For all the proofs of theorems and lemmas, we refer the reader to Appendix~\ref{sec:theory}.

\vspace{-0.125cm}
\subsection{Naive Algorithm}
\label{sss:naive_algorithm}
\noindent Recall that matching is determined by a function $\pi$ from $[1,n]$ to $[1,m]$, and that the cost of {\em any} matching, $\pi$, is given by $C(\pi) = ${avg}$_{i=1}^{i=n}{d(f_i,b_{\pi(i)})}$. The optimal matching is one that minimizes this cost. Let us denote the optimal  matching by $\pi^*$ and the associated cost by $C(\pi^*)$. That is, $\pi^* = \arg\min_{\pi}${avg}$_{i=1}^{i=n}{d(f_i,b_{\pi(i)})}$.
The path smoothness assumption results in the following lemma.\vspace{0.1cm}

\begin{lemma} \label{lem:opt}
\noindent $C(\pi^*) \leq \epsilon$.
\end{lemma}
\noindent A naive brute-force matching algorithm finds the best background match for every foreground frame, \textit{i.e.}, for each $f_i$, compute $d(f_i, b_j)$ $\forall$ $j$ and pick the one that minimizes $d(f_i, b_j)$. The time complexity of this algorithm is $\Theta(mn)$.

\subsection{Speed-Up with Provable Error Bound}
\label{sss:quadratic_speedup}

\noindent We now present a more efficient video matching scheme detailed in Algorithm~\ref{algo:matching} (cf. {\sc Near-Linear}). The main intuition here is that rather than iterating through every foreground frame in search of its match in the background frame sequence, we only consider foreground frames in chosen intervals of $k$. The remaining foreground frames can then be matched based on the two considered foreground frames sandwiching it. To obtain an additional improvement, a similar idea is leveraged in the background video, by making comparisons only in jumps of $k$ background frames.

Due to path smoothness, we know that every foreground frame that got {\em omitted} in consideration is {\em close enough} to a chosen foreground frame, in particular, at most $\frac{k}{2}$ steps away. Furthermore, due to capture completeness, there exists a background frame that is sufficiently close to each chosen frame. Even though we may have skipped the ideal match, the crucial observation here is that we can apply smoothness on the background as well. This intuition can indeed be formalized concretely. Related ideas that enhance this algorithm are reserved for the next subsection. The following theorem formalizes the error bounds.

\begin{algorithm}[tb]
\caption{{\sc Near-Linear}($k$)}
\label{algo:matching}
{\bf Input:} $\bar{f}$, $\bar{b}$, and parameter $k$.\\
{\bf Output:} $\pi$, matched backgrounds $\forall f \in \bar{f}$.\\\vspace{-0.275cm}
\begin{algorithmic}[1]
\STATE Let $S^k_i = \{i: 1\leq i\leq n, i = 0 \mod k\} \cup \{1\} \cup \{n\}$.
\STATE Let $S^k_j = \{j: 1\leq j\leq m, j = 0 \mod k\} \cup \{1\} \cup \{m\}$.
\STATE $\forall i\in S^k_i$, set $\pi(i) = \arg\min_{j\in S^k_j}{d(f_i, b_j)}$.
\STATE $\forall i\notin S^k_i$, let $near(i) = \arg\min_{t\in S^k_i}{|t-i|}$.
\STATE $\forall i\notin S^k_i$, set $\pi(i) = \pi(near(i))$.
\end{algorithmic}
\end{algorithm}

\begin{theorem} \label{thm:matching}
Algorithm {\sc Near-Linear($k$)} runs in $O(\frac{nm}{k^2})$ time and returns $\pi$ such that every foreground matches to a background with additive error of at most $k\delta$, implying $C(\pi) \leq C(\pi^*) + k\delta\leq \epsilon + k\delta$.
\end{theorem}

\noindent Note that we may choose $k$ as any value between $1$ and $\min\{n,m\}$. This allows us to trade-off between efficiency and accuracy parametrically. For example, if we choose $k\approx \sqrt{m}$, we get a near-linear running time. We state a corollary for an alternative parameter choice that admits the $k^2$ speed-up while guaranteeing near-optimality.

\begin{corollary}
For $k=\frac{\epsilon}{\delta}$, {\sc Near-Linear($k$)} guarantees an $O(\epsilon)$ cost, and yet runs $k^2$ times faster.
\end{corollary}
\noindent The effect of varying $k$ on the algorithm's accuracy is empirically analyzed in a systematic manner in $\S$~\ref{sec:simulations}.

\subsection{Clustered Matches}

\noindent We now show that given any foreground frame, if there are {\em strong matches} (defined below) in the background, they tend to be in close temporal proximity of each other. This understanding allows for a more efficient matching algorithm that can exploit a {\em local} constraint. In this section we show the theoretical result based on a realistic randomized model that we introduce, and $\S$~\ref{sec:simulations} corroborates our findings through controlled simulation experiments.

Recall that a background frame $b$ corresponds to a strong match for $f$ if the distance $d(f, b)$ is {\em small}. Let us formalize this notion more crisply here.

\vspace{0.1cm}\noindent\textbf{Strong Match: } We define $b$ to be a strong match for $f$ if $d(f,b)\leq \Psi$. Moreover, given a strong match $\{f,b\}$, we model the measure of distance $d(f,b)$ as a sample from the uniform distribution $[0, \Psi]$.

\begin{algorithm}[tb]
\caption{{\sc Local-Search}(${\bar{f}}, {\bar{b}}$)}
\label{algo:local}
{\bf Input:} $f$, $\bar{b}$, $k$, and $\gamma$.\\
{\bf Output:} Matched background $\forall f \in \bar{f}$.\\\vspace{-0.275cm}
\begin{algorithmic}[1]
\STATE Let $S^k_j = \{j: 1\leq j\leq m, j = 0 \mod k\} \cup \{1\} \cup \{m\}$.
\STATE For $j\in S^k_j$, compute if $b_j$ is a {\em strong match} for $f$.
\STATE $\forall$ $b_j$ that are strong matches, compare $f$ with each $b \in b_x$ where $x : [j-\gamma, j+\gamma]$ to find other strong matches.
\STATE Return all strong matches found.
\end{algorithmic}
\end{algorithm}

\vspace{0.05cm}\noindent Before presenting the main result of this subsection, we first state a few prerequisite results.

\begin{lemma} \label{lem:singleneighbor}
Suppose that $b_i$ is a strong match for a foreground frame $f$. Then, the probability that $b_{i\pm \gamma}$ is also a strong match for $f$ is at least $\frac{\Psi-\gamma\delta}{\Psi}$.
\end{lemma}

\begin{lemma} \label{lem:windowexpectation}
Given a strong match between $f$ and $b_i$, the expected number of strong matches for $f$ in the background sequence range $[b_{i-\gamma},b_{i+\gamma}]$ is at least $2(\gamma+1)(1-\frac{\delta\gamma}{2\Psi}) - 1$.
\end{lemma}

\noindent Recall that capture completeness only gives an upper bound guarantee even for the {\em best} match, specifically a distance of at most $\epsilon$. We therefore consider {\em strong} matches as having a weaker upper bound guarantee, \textit{i.e.}, $\Psi \approx c\epsilon$ (i.e. $\frac{\Psi}{\delta}\gg c$). In such a situation, we wish to interpret Lemma~\ref{lem:windowexpectation} with reasonable parameter values. We present one such choice of parameters in the following theorem and then proceed to describe the algorithm that would guarantee efficiency in terms of a speed-up similar to $k^2$ as before, without the added cost of missing out on strong matches.

\begin{theorem} \label{cor:clustered}
Given a strong match, with a window size $\gamma = \frac{\Psi}{\delta}$, the expected number of strong matches in the entire window of size $(2\gamma+1)$ is at least $\gamma$.
\end{theorem}

\noindent This leads to an algorithm that suggests making $\gamma$ skips in the background frames and whenever encountering a strong match, adopting a local search approach. Theorem~\ref{cor:clustered} therefore suggests finding several strong matches without performing exhaustive search. This procedure is listed for one foreground frame in Algorithm~\ref{algo:local}, and can be extended $\forall f \in \bar{f}$ with strong background matches. For foreground frames with no strong background matches, we propose to rely on foreground tracking (see $\S$~\ref{ssec:foreground_tracking} for details).



%
\section{Computational Framework}
\label{sec:comp_framework}
\begin{figure}[!t]
\begin{center}
\includegraphics[width=1.0\linewidth]{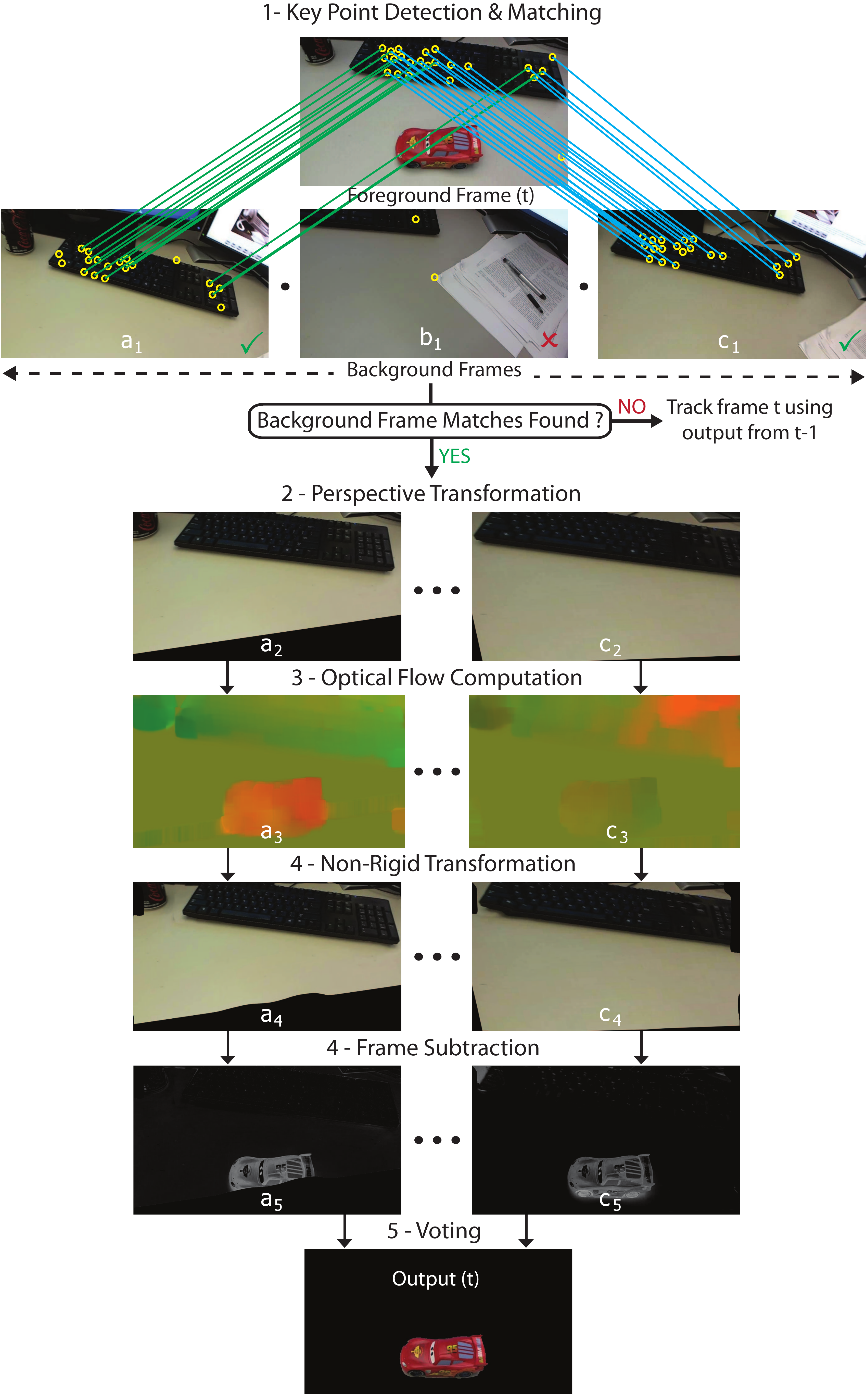}
\end{center}
\caption{Illustration of the different steps in our proposed computational framework.}
\label{fig:framework_col}
\end{figure}

\noindent We now present a foreground extraction framework that exploits our frame matching algorithm as explained in $\S$~\ref{sec:Spatio-Temporal-Frame-Matching}. The main steps of our framework are illustrated in Figure~\ref{fig:framework_col}. There are two important insights to be noted here:

\vspace{0.1cm}\noindent ${\bullet}$ While our scene might have depth, we show that using projective transform followed by non-rigid frame warping generally works sufficiently well for aligning the background and foreground frames.\vspace{0.1cm}

\vspace{0.1cm}\noindent ${\bullet}$ Furthermore, the warping error obtained for a foreground frame varies over different matched background frames. This allows us to integrate over multiple noisy warping hypotheses to obtain a robust inference about the foreground.\vspace{0.1cm}

\vspace{0.15cm}\noindent Recall that approximating scene geometry without using the complete $3$-D scene structure has previously been explored~\cite{shum1998construction}~\cite{seitz1995complete}. The novelty of our framework however lies in applying this idea to obtain \textit{multiple} noisy estimates of foreground, and to then \textit{fuse} these multiple hypotheses to obtain a final robust inference. Our results show that combining these two steps is crucial for efficient background subtraction of videos tracing substantially different camera paths. We now explain the steps of our framework.


\subsection{Key-Point Detection \& Frame Matching}
\label{ss:key_point_detection}
\noindent For each foreground and background frame, we compute key-points descriptors using the SURF pipeline~\cite{bay2006surf}, and find matches between them using the RANSAC algorithm~\cite{fischler1981random}.

We define the distance between a pair of background and foreground frames, $d(f,b)$, as the reciprocal of the number of key-points matches found between them\footnote{While there can be several measures of frame distance, the number of key-point matches has been shown to be a reasonable choice~\cite{liu2011sift}.}, and search for corresponding background and foreground frames with strong matches using our proposed method explained in $\S$~\ref{sec:Spatio-Temporal-Frame-Matching}.

\subsection{Perspective Transformation}
\noindent For each foreground frame $f$, given a strong match $b \in \bar{b}$, we compute a perspective transform $\zeta_{f,b}$ between them~\cite{hartley2000multiple} such that $b = \zeta_{f,b} f$. Using $\zeta_{f,b}$, we warp $b$ to align it with the given foreground frame~\cite{glasbey1998review}, and generate the transformed background frame $b^{'}$, such that $b^{'} = \zeta^{-1}_{f,b} b$.

\subsection{Non-Rigid Transformation}
\noindent Using perspective transformation, most parts of $f$ and $b^{'}$ get well-aligned with each other. However, there still may be parts where this alignment is not accurate. This disparity is possible because of the planar assumption that perspective transformation makes regarding the scene. However, for our problem this assumption is not necessarily true. To this end, we first compute the optical flow between $f$ and $b^{'}$~\cite{farneback2000fast}. We then apply the computed flow-fields to warp $b^{'}$ onto $f$, to find the non-rigidly transformed background frame $b^{''}$~\cite{glasbey1998review}.\vspace{-0.1cm}

\subsection{Frame Subtraction}\vspace{-0.1cm}
\noindent For each pair of $f$ and $b^{''}$, we perform a per-pixel frame subtraction between their corresponding color channels\footnote{While color as a measure of pixel difference worked well for us, one could also use more robust pixel features instead \textit{e.g.}, dense SURF~\cite{bay2006surf} \textit{etc}.}. The subtracted pixels greater than a certain threshold $\tau_{s}$ are classified as belonging to the foreground scene, and used to obtain the foreground mask $\mathcal{F}$, \textit{i.e.}
\begin{equation}
\small{
  \mathcal{F}(x,y) = \begin{cases}
    1 & \text{if $f(x,y)-b^{''}(x,y) > \tau_{s}$}\\
    0 & \text{otherwise}.
  \end{cases}
  }
\end{equation}

\subsection{Per-Pixel Voting}
\noindent The frame subtraction operation gives multiple candidate hypotheses $\mathcal{F}^{i}(x,y)$ characterizing whether a pixel in the foreground frame belongs to the foreground scene or not. For each of the foreground frame pixels we count the number of times it is classified as belonging to the foreground scene. If this fraction is more than a certain threshold $\tau_{v}$, it is finally classified as a foreground scene pixel $\mathcal{R}(x,y)$, \textit{i.e.},
\begin{equation}
\small{
  \mathcal{R}(x,y) = \begin{cases}
    1 & \text{if avg$_{i}(\mathcal{F}^{i}(x,y)) > \tau_{v}$}\\
    0 & \text{otherwise}.
  \end{cases}
  }
\end{equation}

\begin{algorithm}[tb]
\caption{{\sc Voting-Scheme}(${\bar{f}}, {\bar{b}}$)}
\label{algo:voting}
{\bf Input:} $f\in \bar{f}$, and a set of strong match background frames $b_{s}(f) \subseteq \{b_1, b_2, \ldots, b_m\}$. Threshold parameter $t$.\\
{\bf Output:} Background subtracted image corresponding to $f$.\\\vspace{-0.35cm}
\begin{algorithmic}[1]
\STATE Let $|b_{s}(f)| = r$.
\STATE For every pixel that is labeled as foreground on at least $\tau_{v}$ strong matches, mark it as foreground in the result.
\STATE All other pixels are labeled as background.
\end{algorithmic}
\end{algorithm}

\noindent This procedure is listed in Algorithm~\ref{algo:voting} (cf. {\sc Voting-Scheme}). Note that the likelihood of classifying a pixel correctly increases exponentially with the increase in the number of strong matches. Formally, assume that for any pixel $p \in \mathcal{F}$, we have the following two properties:

\vspace{0.1cm}\noindent ${\bullet}$ If the pixel is from foreground, then it is reflected as so with probability $\geq p_1$, and

\vspace{0.1cm}\noindent ${\bullet}$ If the pixel is from background, then it is reflected as so with probability $\geq p_2$ where $p_1+p_2>1$.

\vspace{0.1cm} \noindent Then the following lower-bound holds on the probability of correctly classifying $p$.
\begin{theorem} \label{lem:voting_lemma}
Given $r$ strong matches, and a voting threshold $t$ in {\sc Voting-Scheme} in the open interval $(1-p_2, p_1)$, the probability that any pixel is classified correctly is at least $1-e^{-O(r)}$.
\end{theorem}

\noindent Intuitively, Theorem~\ref{lem:voting_lemma} suggests that for each foreground frame, we only need a few background frames with strong matches to perform foreground extraction robustly.

\subsection{Selective Foreground Tracking}
\label{ssec:foreground_tracking}
\noindent There might be foreground frames for which there are no strong background matches to be found. For such frames, we propose to track the latest detected foreground from their previous frame that did have strong background matches. For this work, we found Mean-Shift tracking~\cite{comaniciu2000real} to be an efficeint choice for tracking the foreground when needed.

Since for any foreground frame, tracking the foreground pixels is only a constant time operation, it does not effect the overall complexity of the algorithms described in $\S$~\ref{sec:Spatio-Temporal-Frame-Matching}.
\section{Simulation Analysis}
\label{sec:simulations}
\noindent Since the paths traced by handheld cameras can be unconstrained, it is important to analyze how the accuracy of our framework varies as a function of the difference between the cameras paths traced by the background and foreground videos. We now present a set of controlled simulation experiments to analyze this question in a systematic manner.

\subsection{Video Data Generation}
\label{ss_data_generation}
\noindent We used a $3$-D animation software (Blender) to design a desktop scene (Figure~\ref{fig:problem_statement_fig}) where we modeled the background camera path as a circle parameterized by a set of key points. To generate the foreground camera path, we randomly perturbed these key-points by different amounts, and fitted a spline curve to them. We quantified the added perturbation added as a fraction of its radius. For each perturbation amount, we generated $10$ sets of foreground paths.

\begin{figure}[!t]
\begin{center}$
\begin{array}{cc}
\includegraphics[width=0.5\linewidth]{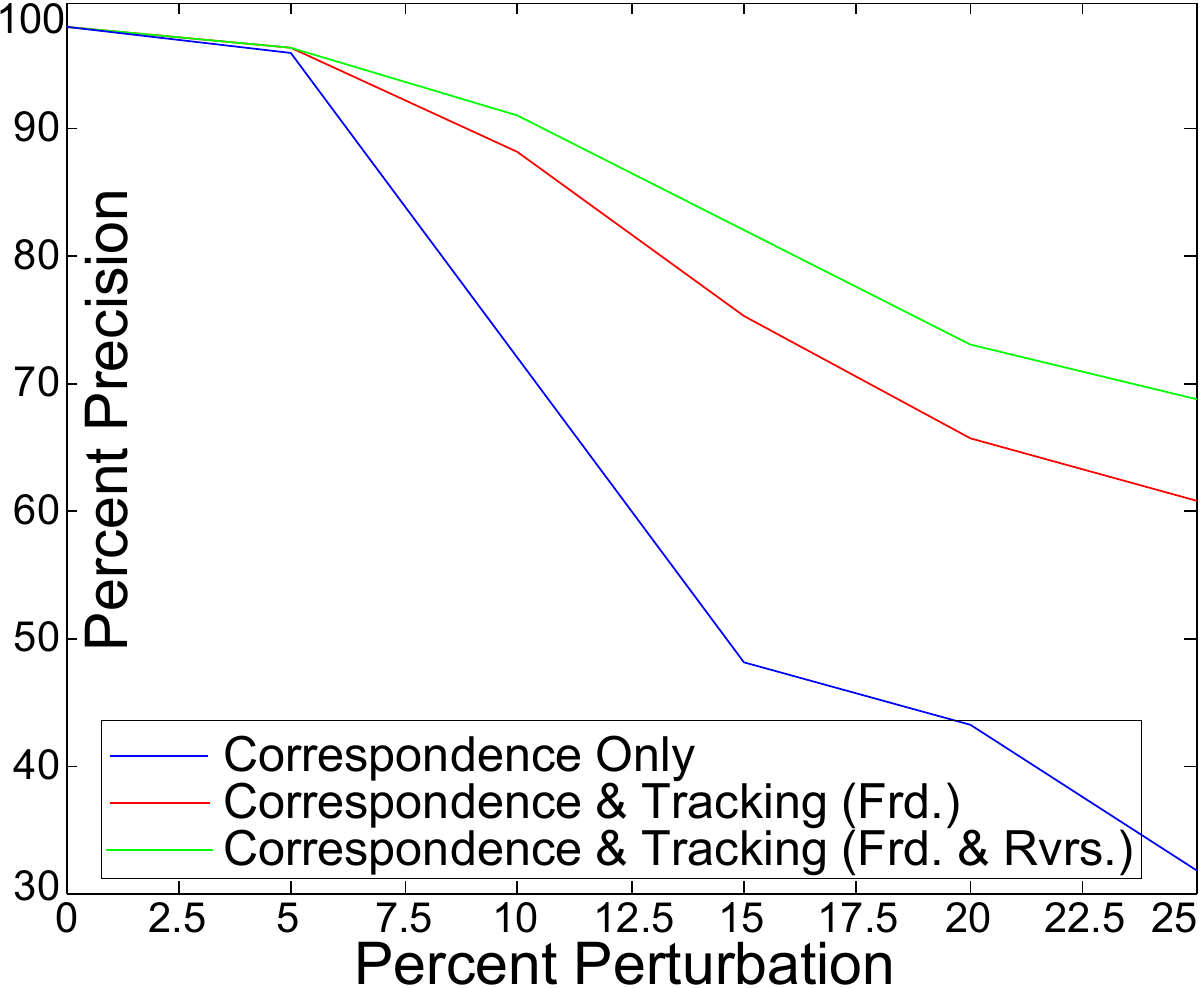}
\includegraphics[width=0.5\linewidth]{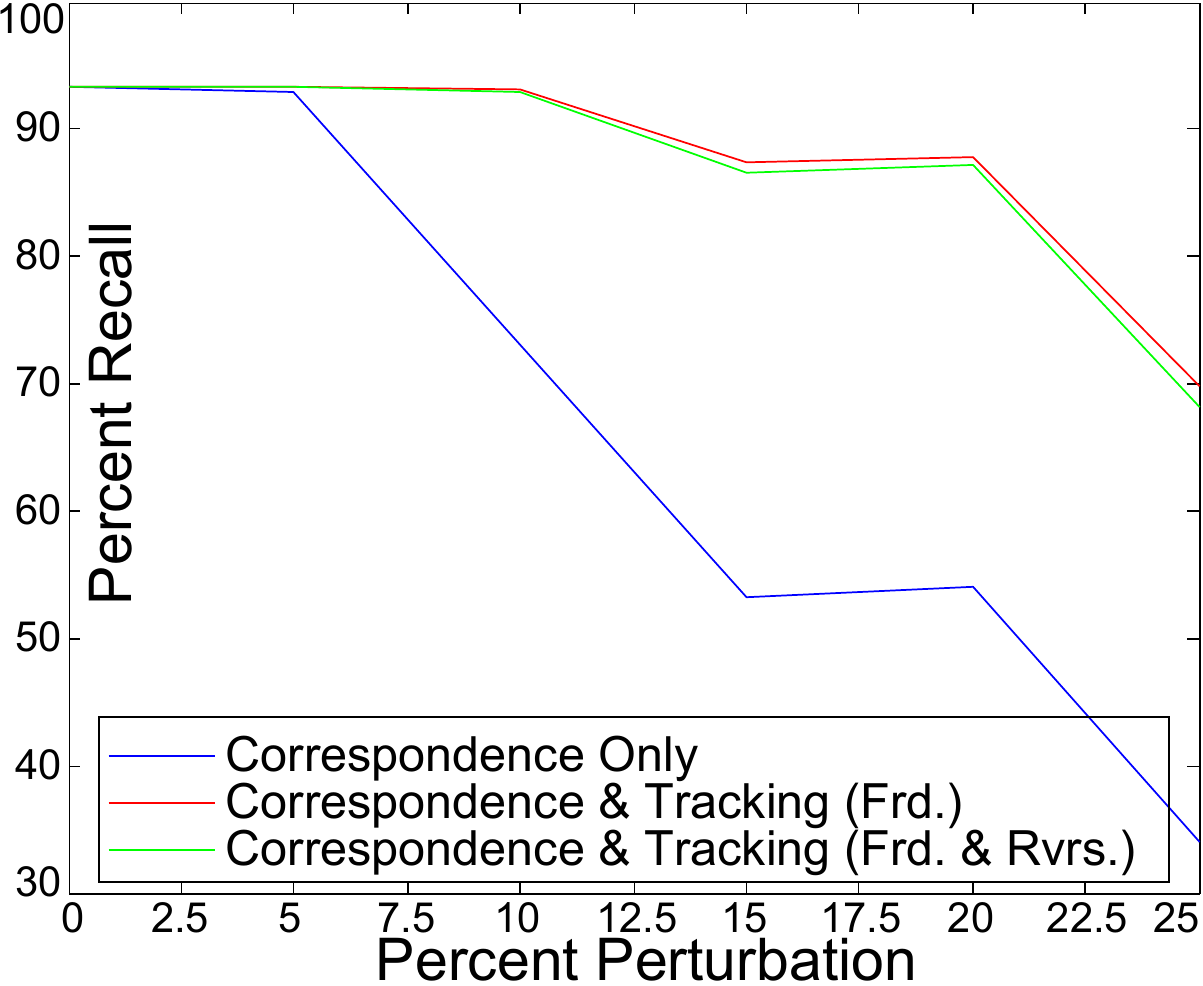}\\
\includegraphics[width=0.5\linewidth]{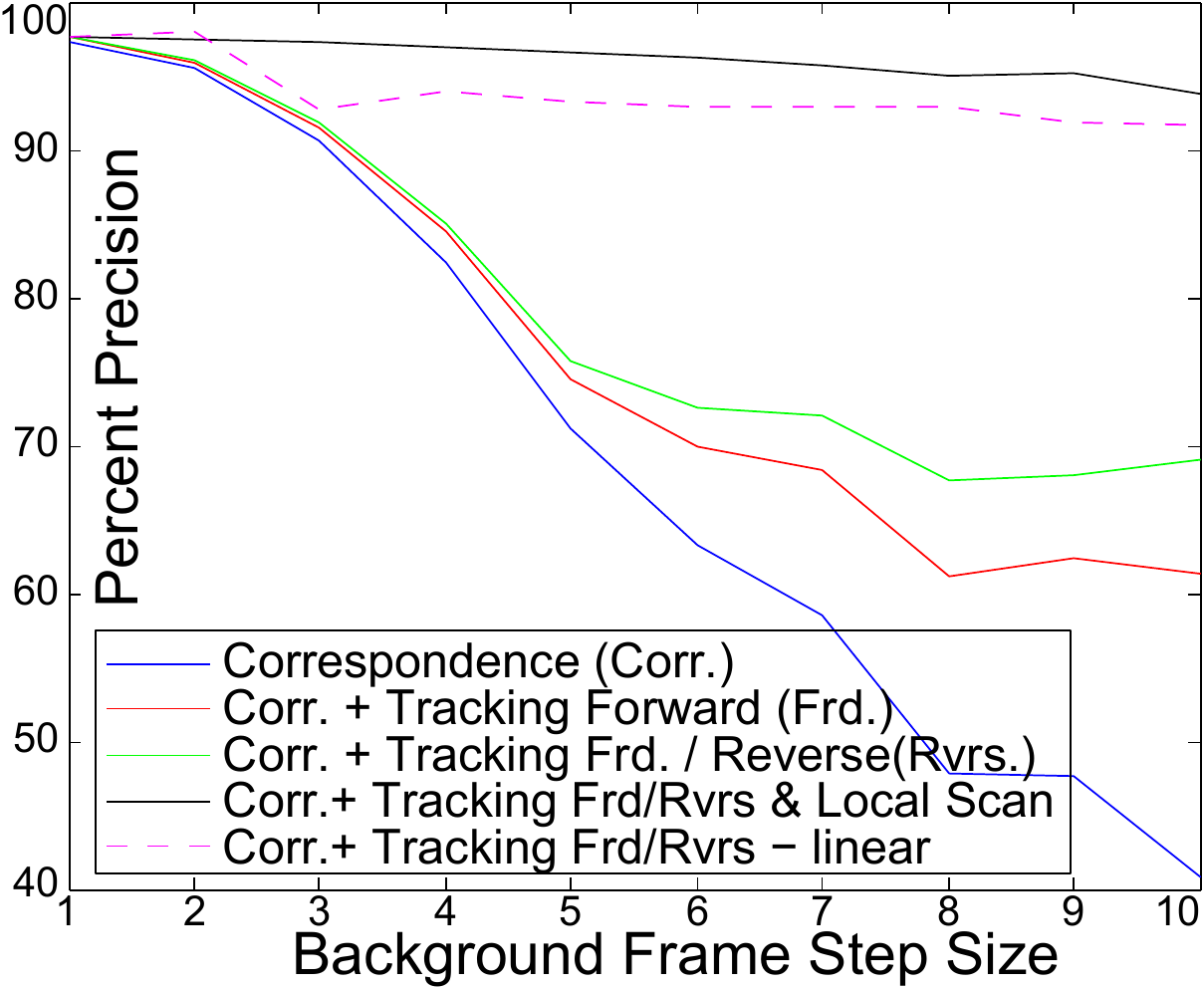}
\includegraphics[width=0.5\linewidth]{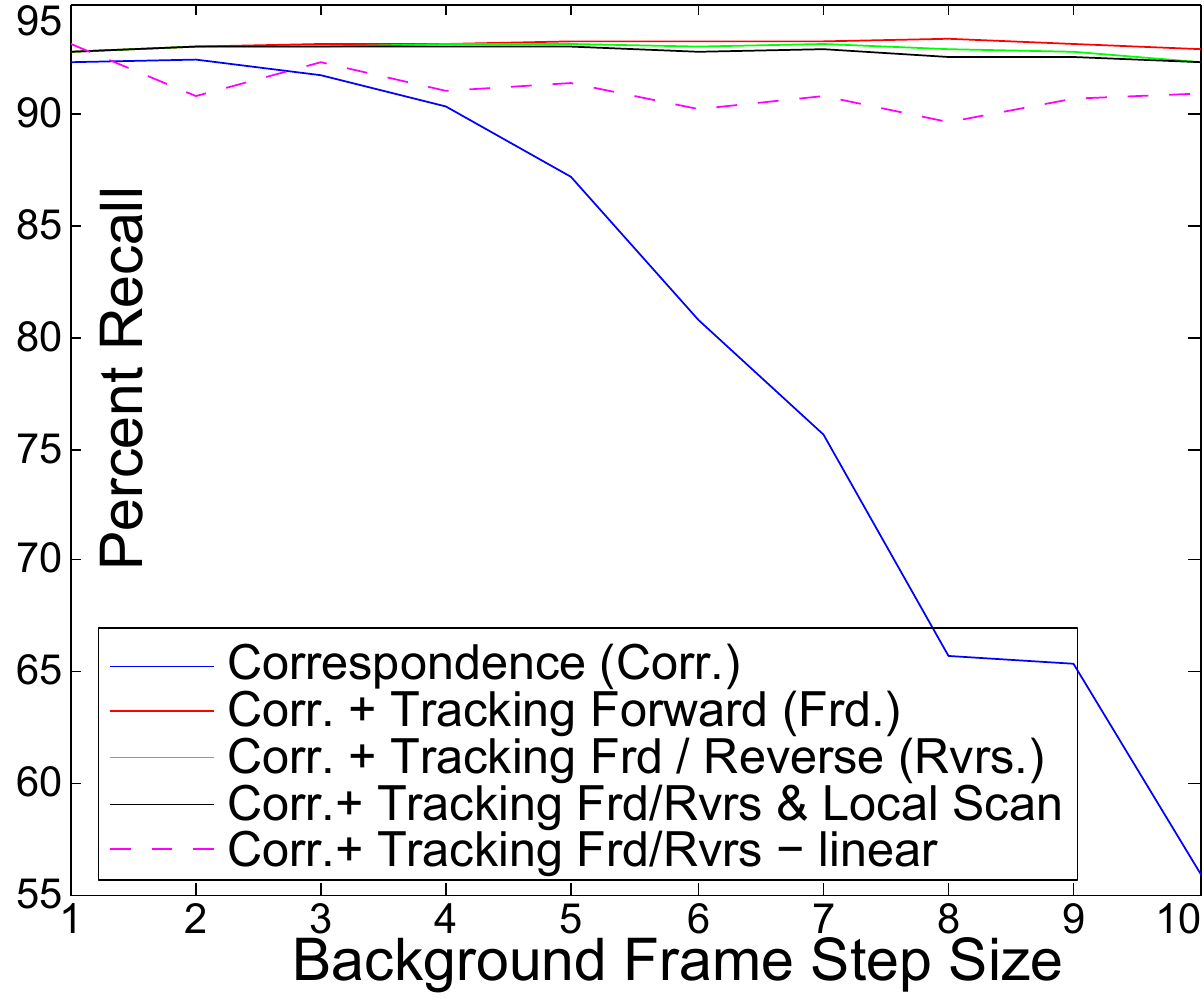}\\
\end{array}$
\end{center}
\caption{Precision (top left) and Recall (top right) plots with increasing amounts of camera path perturbation. Blue graph shows results obtained by only using frame matching (corr.), while green and red graphs represent also using forward (frd.) and reverse (rvrs.) tracking of foreground pixels in frames that do not have strong matches in the background video. Precision (bottom left) and Recall (bottom right) are shown as a function of background frame step-size. The black and purple graphs use local search in the background, and both background and foreground videos.}
\label{fig:sim_results}
\end{figure}

\subsection{Simulation Experiments}
\noindent We ran three main experiments on our generated data.

\vspace{0.1cm}\noindent \textbf{Experiment 1:} In this experiment (Figure~\ref{fig:sim_results}, top row), we analyzed our precision and recall as a function of foreground path perturbation\footnote{As we know the ground truth, we can compute the precision and recall based on their standard definitions.}. Here, we searched for strong matches in a naive manner ($\S$~\ref{sss:naive_algorithm}).

\vspace{0.1cm}\noindent \textbf{Experiment 2:} For our second experiment, we analyzed how strong background frame matches are temporally distributed. We performed this experiment for different amounts of foreground path perturbations (Figure~\ref{fig:sim_temporal_clustering_25_090_10}).

\vspace{0.1cm}\noindent \textbf{Experiment 3:} Finally, for our third experiment (Figure~\ref{fig:sim_results}, bottom row), we analyzed for a given path perturbation, how does our performance vary as we search for strong background frame matches in a more greedy way ($\S$~\ref{sss:quadratic_speedup}).

\subsection{Main Findings}
\label{ss_data_generation}
\noindent We now report our main findings from these experiments.

\subsubsection{Matching Alone is Not Enough}
\label{ss:mathing_not enough}
\noindent When purely relying on finding a strong matching between the foreground and the background frames, the performance of our framework degrades steeply (blue plots in Figure~\ref{fig:sim_results}, top row). This is because for larger disparity between the background and foreground camera paths, there may be no strong matches available, \textit{i.e.}, capture completeness ($\S$~\ref{sssec:capture_completeness}) may not be true for all foreground frames. However, as can be seen from the red plots in the top row of Figure~\ref{fig:sim_results}, adaptively switching between background subtraction of registered frames, and tracking the foreground pixels based on their latest available location information substantially improves the performance.\vspace{-0.2cm}

\subsubsection{Temporal Smoothing Improves Accuracy}
\label{ss:frd_brd_tracking}
\noindent Performing foreground tracking for frames with no matching background can be improved by combining a forward and a reverse tracking pass (Figure~\ref{fig:sim_results}, top row). This is because for the foreground frames for which we do have strong background matches, the noisy pixels appear and disappear erratically, while the foreground pixels are much more consistent over frames. Therefore, fusing information of foreground pixels obtained by tracking them in the forward and reverse direction results in accuracy improvement at only a constant additional cost in time complexity.\vspace{-0.2cm}
\subsubsection{Matches are Temporally Clustered}
\begin{figure}[t]
\begin{center}
   \includegraphics[width=1.0\linewidth]{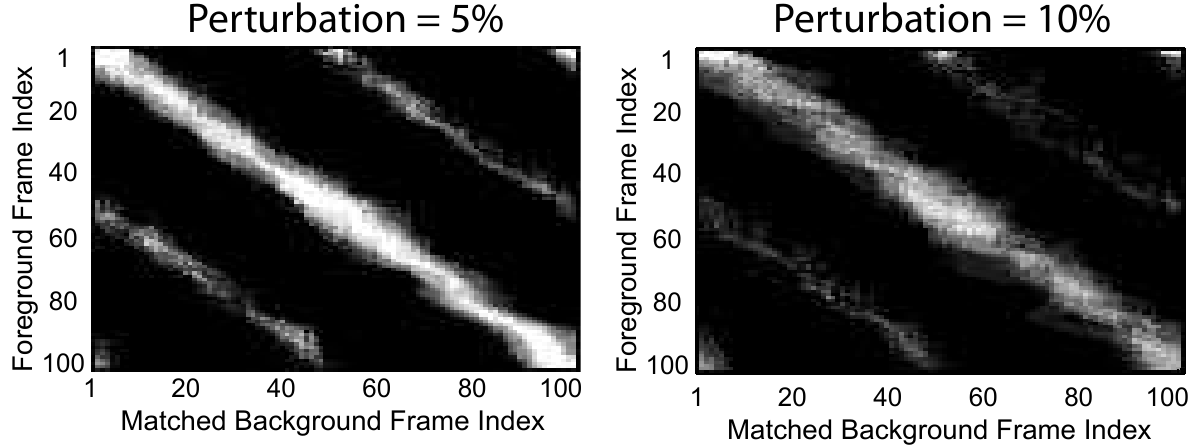}
\end{center}
   \caption{For each foreground frame, we show how likely a background frame is to be a \textit{strong match}. Here white implies high likelihood. Results for $5\%$ $\&$ $10\%$ perturbation are shown. Greater perturbation amounts follow similar trends.}
\label{fig:sim_temporal_clustering_25_090_10}
\end{figure}

\noindent We analyzed how strong background frame matches are temporally distributed in the background video. This analysis corroborates our theoretical result of Theorem~\ref{cor:clustered} which suggested that several strong matches are likely to be clustered together. This can be observed from Figure~\ref{fig:sim_temporal_clustering_25_090_10}, where light regions show pairs of foreground and background frames that form a strong match (averaged over $10$ independent trials). As can be seen, light regions cluster together, lending empirical evidence to our theoretical result.
\subsubsection{Selective Coarse-to-Fine Frame Search Works}
\noindent To find strong matches for a foreground frame, it is an efficient and accurate strategy to first sparsely search a well-distributed set of background frames, followed by a more dense search around background frames that match well with the given foreground frame. This is in accordance with Algorithm~\ref{algo:local}, and is empirically corroborated by the results shown in Figure~\ref{fig:sim_results} (bottom row).

As can be seen, with the forward and reverse tracking, and the local search principle incorporated (the dashed purple graph), the algorithm's performance (both for precision and recall) is nearly unaffected even as the step size is increased. Specifically, the precision and recall are both above $90\%$ for step sizes all the way up to $k=10$. Note that as the step size increases, it significantly outperforms all approaches that do not leverage local search. Further, it comes quite close to the black graph (the overall best) despite being an additional $k$ times faster than it. Note that for these experiments, the number of video frames was $100$ each. This implies that our framework can achieve greater than $90\%$ accuracy while taking time only linear in the lengths of the background and foreground videos.\vspace{-0.2cm}


\section{Experiments \& Results}
\label{sec:experiments}
\begin{figure}[!t]
\begin{center}
   \includegraphics[width = 1.0\linewidth]{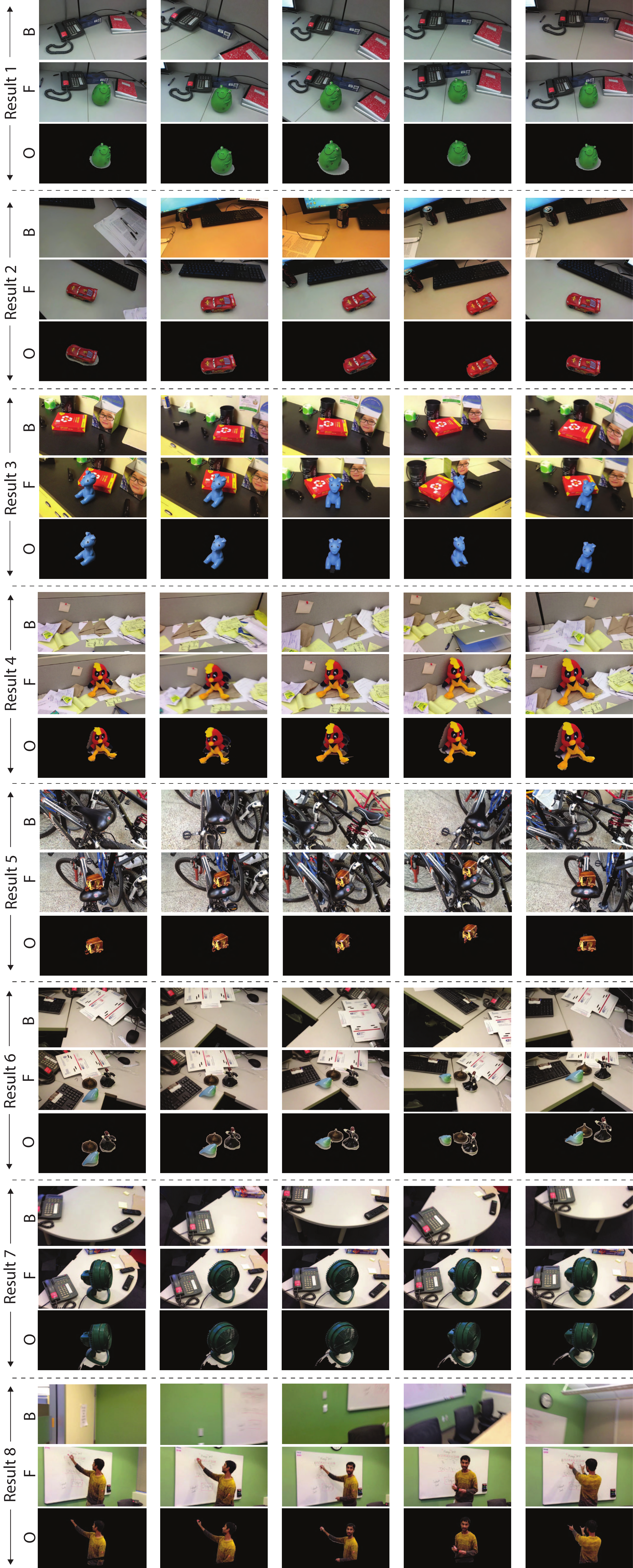}
\end{center}
   \caption{Results capturing various experiment parameters are presented. Here B, F, and O denote background, foreground, and output videos.}
\label{fig:results}
\end{figure}

\noindent We now present results of our framework tested on real videos taken by multiple users.

\subsection{Explored Test Scenarios}
\noindent We performed a set of experiments designed to test different challenges for handheld cameras. Sample video frames and output from for these are presented in Figure~\ref{fig:results}. The full videos are provided in the supplementary material.\vspace{0.1cm}

\noindent\textbf{Foreground Object Geometry}: The first five results in figure~\ref{fig:results} are presented in the order of increasing foreground complexity. As shown, our framework performs well for a range of object geometries including mostly convex (result 1) to quite non-convex (result 5).\vspace{0.1cm}

\noindent\textbf{Indoors Versus Outdoors}: In Figure~\ref{fig:results}, results $1$ to $4$ were captured indoors, while result $5$ was captured outdoors in daylight. Even within the indoor environment the auto white-balancing feature of mobile cameras can result in substantial variation in the global illumination of the video. Our framework is able to handle these variations quite well.\vspace{0.1cm}

\noindent\textbf{Multiple Foreground Objects}: Our framework is agnostic to the number of foreground objects present in the video. This is shown in result $6$ of Figure~\ref{fig:results}, where we were able to successfully extract all three foreground objects.\vspace{0.1cm}

\noindent\textbf{Moving Foreground Objects}: We captured videos for both simple and articulated moving objects, shown in the last two results of Figure~\ref{fig:results}. Not only do we preform well for the limited movement of the table fan, we were also able to extract the articulated human motion. This result is a first step for automatic green-screening of user-generated videos.

\begin{figure}[t]
\begin{center}
   \includegraphics[width = 0.925\linewidth]{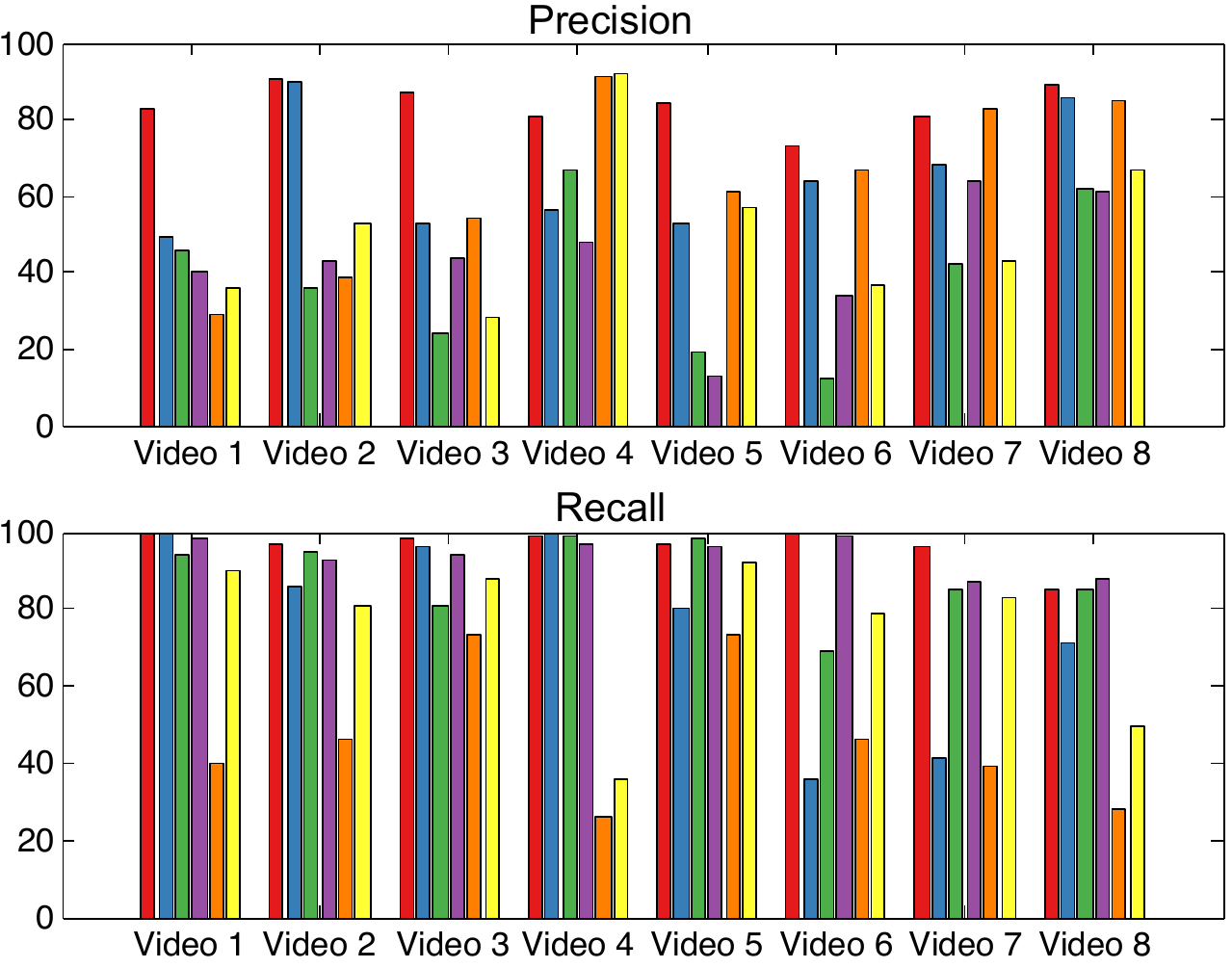}
\end{center}
   \caption{Precision and Recall plots of our proposed method (red) and five comparative approaches.}
\label{fig:comp_results}
\end{figure}

\subsection{Comparative Results}
\label{ss:comparative_results}

\noindent We compared our algorithm with five alternate foreground extraction methods. These methods are explained below.

\noindent ${\bullet}$ In the first comparison (blue plot), we used a Gaussian Mixture model (GMM) to learn the global background appearance which was used for background subtraction~\cite{stauffer1999adaptive}.

\noindent ${\bullet}$ In the second comparison (green plot), we manually provided a foreground mask for the first frame of the foreground video to learn a foreground GMM. This model was used to subtract background from all the foreground frames.

\noindent ${\bullet}$ Our third comparison (green plot) was with the previously proposed approach of video matching~\cite{sand2004video}, and highlights its limitations due to its strong assumptions about the foreground and background camera paths being the same.

\noindent ${\bullet}$ Our fourth comparison (orange plot) was with a perceptual saliency based approach for the foreground frames~\cite{itti1998model}.

\noindent ${\bullet}$ Our fifth and final comparison (yellow plot) was with a multi-frame co-segmentation based approach~\cite{meng2010object}.

\subsubsection{Precision \& Recall}
\noindent We manually extracted the foreground object(s) in every $10^{\textrm{th}}$ frame of the test videos presented in Figure~\ref{fig:results}. Using this, we computed the number of true/false positives and negatives. We report the average precision and recall for our method in Figure~\ref{fig:comp_results} (red plot). As shown, our approach gives better recall as well as comprehensively outperforms previous approaches on precision.

\subsubsection{Time Comparisons}

\noindent Clock times taken by three of the five considered algorithms on the test videos are shown in Table~\ref{tbl:clock_times}. Note that the remaining two algorithms from Figure~\ref{fig:comp_results} are derivatives of the three considered methods in Table~\ref{tbl:clock_times}, with times similar to the first two rows of the table. As can be observed, our proposed method takes the least amount of processing time while generating superior results compared to all the considered alternatives.

\begin{table}[b]
\begin{small}
{
\begin{center}
\begin{tabular}{|l||c|c|c|c|c|c|c|c|c|}
  \hline
  & $V_{1}$ & $V_{2}$ & $V_{3}$ & $V_{4}$ & $V_{5}$ & $V_{6}$ & $V_{7}$ & $V_{8}$\\
  \hline
   \hline
  $A_{1}$ & 41.7 & 11.8 & 19.1 & 5.8 & 83.6 & 42.5 & 18.6 & 47.1\\
  \hline
  $A_{2}$ & 10.6 & 6.8 & 11.7 & 4.7 & 5.9 & 5.6 & 8.7 & 10.6\\
  \hline
  $A_{3}$ & \textbf{8.1} & \textbf{4.3} & \textbf{8.7} & \textbf{2.9} & \textbf{5.2} & \textbf{5.3} & \textbf{5.7} & \textbf{9.5}\\
  \hline
  \end{tabular}
\end{center}
\vspace{-0.1cm}\caption{Algorithms' clock times in minutes. Here A$1$, A$2$, and A$3$ represent GMMs based modeling, saliency based extraction, and proposed method.}
\label{tbl:clock_times}\vspace{-0.4cm}
}
\end{small}
\end{table} 
\section{Conclusions \& Future Work}
\noindent In this paper, we identified a novel instance of the background subtraction problem focusing on extracting near-field foreground objects captured using handheld devices. We posed our problem as efficient matching of foreground and background videos tracing significantly different camera trajectories. We proposed an approximate algorithm for this problem with provable theoretical guarantees that admit a significant computational speed-up. Using insights from our theoretical model, we proposed an end-to-end computational framework, and presented controlled simulation experiments to thoroughly analyze its object-extraction accuracy. Moreover, we presented results of our framework on multiple hand-held videos, and compared them with five alternate approaches. Going forward, we plan to also learn the shape and appearance of the extracted foreground object(s). We expect this to allow us to improve our accuracy further, while requiring even fewer matches.

\appendix
\section{Theoretical Analyses} \label{sec:theory}

\noindent In the following we present the detailed proofs for the theorems and lemmas we presented in our paper.\\

\noindent \textbf{Lemma 3.1} $C(\pi^*) \leq \epsilon$.\\

\noindent \textbf{Proof:} Follows from the path smoothness assumption.\\

\noindent \textbf{Theorem 3.2}: Algorithm {\sc Near-Linear($k$)} runs in $O(\frac{nm}{k^2})$ time and returns $\pi$ such that every foreground matches to a background with additive error of at most $k\delta$, i.e. $C(\pi) \leq C(\pi^*) + k\delta\leq \epsilon + k\delta$.\\

\noindent \textbf{Proof}: The proof follows in the following two steps. 

\vspace{0.1cm}\noindent \textbf{Step 1}: We first show the bound $\forall$ $f_i$ where $i\in S^k_i$.  Let $j_{opt} = \arg\min d(f_i, b_{j_{opt}})$.  In algorithm {\sc Near-Linear($k$)}, let the returned value for $\pi(i) = j_{alg}$. Let $near(j_{opt}) = \arg\min_{t\in S^k_j}{|t-j_{opt}|}$. It follows that $|near(j_{opt}) - j_{opt}|$ is at most $k/2$, since $S^k_j$ has all multiples of $k$ as well as $1$ and $m$. Therefore by triangular inequality, $d(f_i, b_{near(j_{opt})}) \leq d(f_i, b_{j_{opt}}) + d(b_{j_{opt}}, b_{near(j_{opt})})$. This is at most $d(f_i, b_{j_{opt}} + \delta\cdot|near(j_{opt}) - j_{opt}|$ by the continuous capture assumption and the triangle inequality (since $d(b_x, b_{x+1})\leq \delta$ for all $x$). It follows that $d(f_i, b_{near(j_{opt})}) \leq d(f_i, b_{j_{opt}}) + k\delta/2$. Notice that $d(f_i, b_{j_{alg}})\leq d(f_i, b_{near(j_{opt})})$, because $near(j_{opt})\in S^k_j$ and the algorithm picked $\pi(i) = j_{alg}$ as $\arg\min_{j\in S^k_j}{d(f_i, b_j)}$. Combining, we get $d(f_i, b_{j_{alg}})\leq d(f_i, b_{j_{opt}}) + k\delta/2$. Furthermore, we know by the complete capture assumption, that $d(f_i, b_{j_{opt}})\leq \epsilon$. It follows that $d(f_i, b_{j_{alg}})\leq \epsilon + k\delta/2$.

\vspace{0.1cm}\noindent \textbf{Step 2}: We now consider the case when $i\notin S^{k}_{i}$. Let $near(i) = \arg\min_{t\in S^k_i}{|t-i|}$. It follows that $|i - near(i)|\leq k/2$, because $S^k_i$ contains all multiples of $k$ between $1$ and $n$, and also includes $\{1\}$ $\{n\}$. As in step $1$, let the returned value for $\pi(i)$ be $j_{alg}$. Therefore, by triangle inequality $d(f_i, b_{j_{alg}})\leq d(f_{near(i)}, b_{j_{alg}}) + \delta\cdot|i - near(i)|$. This is because using the triangle inequality and the continuous capture assumption, $d(b_x, b_y)\leq \delta\cdot|x-y|$. Since $j_{alg}$ is picked as $\pi(near(i)$, it follows that $d(f_{near(i)}), b_{j_{alg}})\leq d(f_i, b_{j_{opt}}) + k/\delta2$. Combining, we get $d(f_i, b_{j_{alg}})\leq d(f_i, b_{j_{opt}}) + k\delta/2 + \delta\cdot|i - near(i)|\leq d(f_i, b_{j_{opt}}) + k\delta/2 + k\delta/2 = d(f_i, b_{j_{opt}}) + k\delta$. As before, $d(f_i, b_{j_{opt}})\leq \epsilon$ by the complete capture assumption. It follows that $d(f_i, b_{j_{alg}})\leq \epsilon + k\delta$. Averaging over all $i$, this implies that $C(\pi) \leq C(\pi^*) + k\delta\leq \epsilon + k\delta$.

\vspace{0.125cm}\noindent Making use of the above two cases, we now show the time complexity analysis of the algorithm {\sc Near-Linear($k$)}. Notice that $|S^k_i| \leq \lfloor n/k \rfloor + 2$ since $S^k_i$ only has multiples of $k$ in $[1, n]$ and $1, n$. Similarly, $|S^k_j| \leq \lfloor m/k \rfloor + 2$. Since the algorithm only computes distances between pairs $f_i$ and $b_j$ where $i\in S^k_i$ and $j\in S^k_j$, and the other computations are just finding $near(i)$ for $i\notin S^k_i$, the total computation time is $O(|S^k_i|\cdot |S^j_j|)$, which is $O(mn/k^2)$. This completes the proof.\\

\noindent \textbf{Corollary 3.3} For $k=\frac{\epsilon}{\delta}$, {\sc Near-Linear($k$)} guarantees an $O(\epsilon)$ cost as the optimal algorithm, but runs $k^2$ times faster.\\

\noindent \textbf{Proof}: Recall that the optimal guarantees only a cost of $\epsilon$. Now, using Theorem $3.2$ plugging in $k=\frac{\epsilon}{\delta}$ gives a cost for algorithm {\sc Near-Linear} equal to $\epsilon + k\delta = 2\epsilon$. Further, the algorithm running time is $mn/k^2$ or $\frac{mn\delta^2}{\epsilon^2}$. Notice that for $\epsilon\gg \delta$ this is a significant improvement on running time at almost no cost on the solution guarantee.\\

\noindent \textbf{Lemma 3.4} Suppose that $b_i$ is a good match for a foreground frame $f$. Then, the probability that $b_{i\pm \gamma}$ is also a good match for $f$ is at least $\frac{\Psi-\gamma\delta}{\Psi}$.\\

\noindent \textbf{Proof}: Since $b_i$ is  good match for $f$, we know that $d(f, b_i)$ is at most $\Psi$ and further is a random variable uniformly distributed in $[0,\Psi]$. Recall by the continuous capture characteristic and the triangle inequality of the distance metric, we know that $d(b_i, b_{i+\gamma})\leq \gamma\delta$. So applying triangle inequality again, we have $d(f, b_{i+\gamma})\leq d(f,b_i) + d(b_i,b_{i+\gamma})$.

Therefore, we know that $d(f,b_{i+\gamma})$ is a random variable that is dominated from above by a random variable that is uniformly distributed in $[\gamma\delta, \Psi+\gamma\delta]$. The probability that such a sample is at most $\Psi$ is at least $\frac{\max\{\Psi-\gamma\delta, 0\}}{\Psi}$ (we drop the max in the lemma statement as if the quantity becomes negative, the lemma is vacuously true). Therefore, this is the probability that $b_{i+\gamma}$ is also a good match for $f$. The exact same proof carries over for $b_{i-\gamma}$ as well.\\

\begin{figure*}[!h]
  \centering

\label{figur:1}\includegraphics[width=88mm]{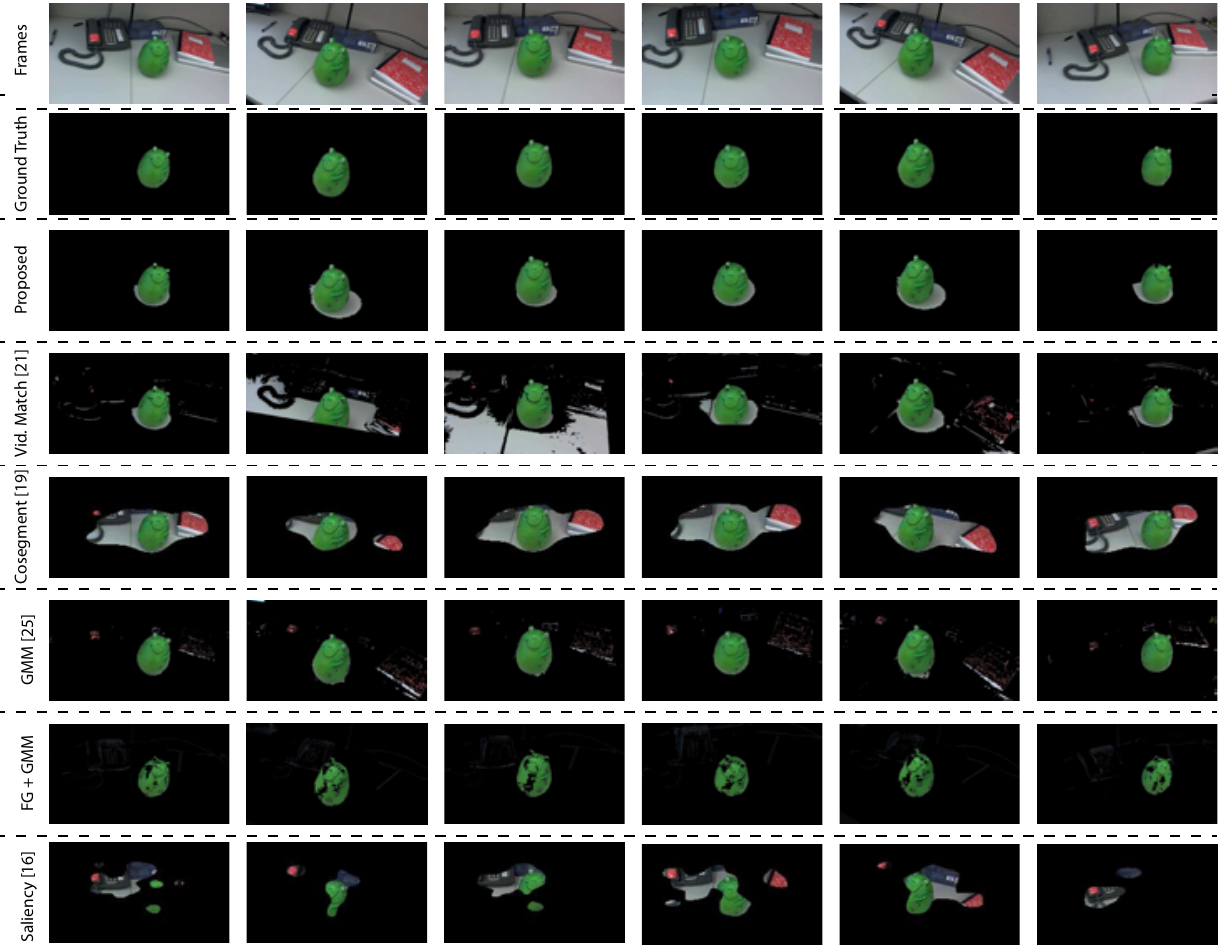}
\label{figur:2}\includegraphics[width=88mm]{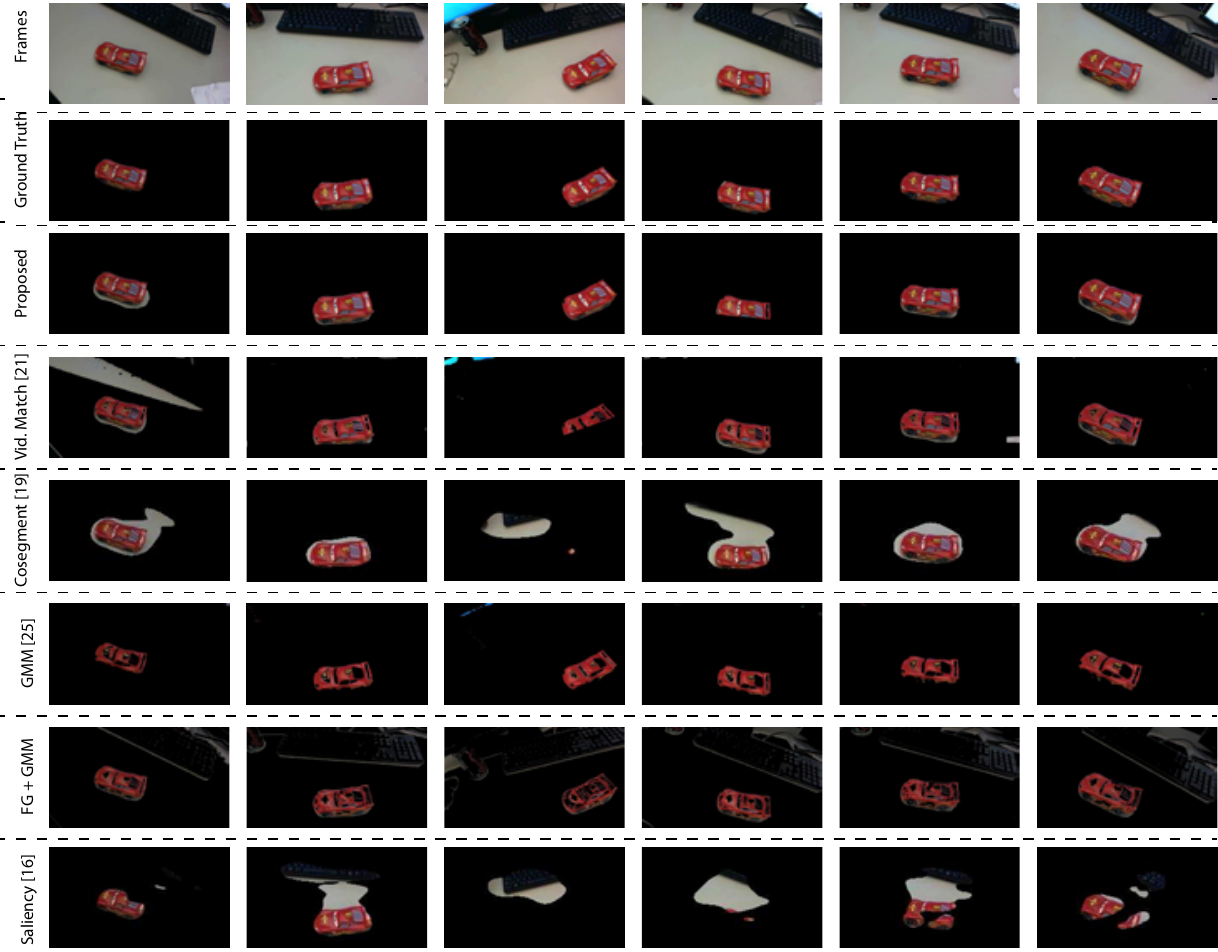}
 --------------------------------------------------------------------------------------------------------------------------------------------
  \\
\label{figur:3}\includegraphics[width=88mm]{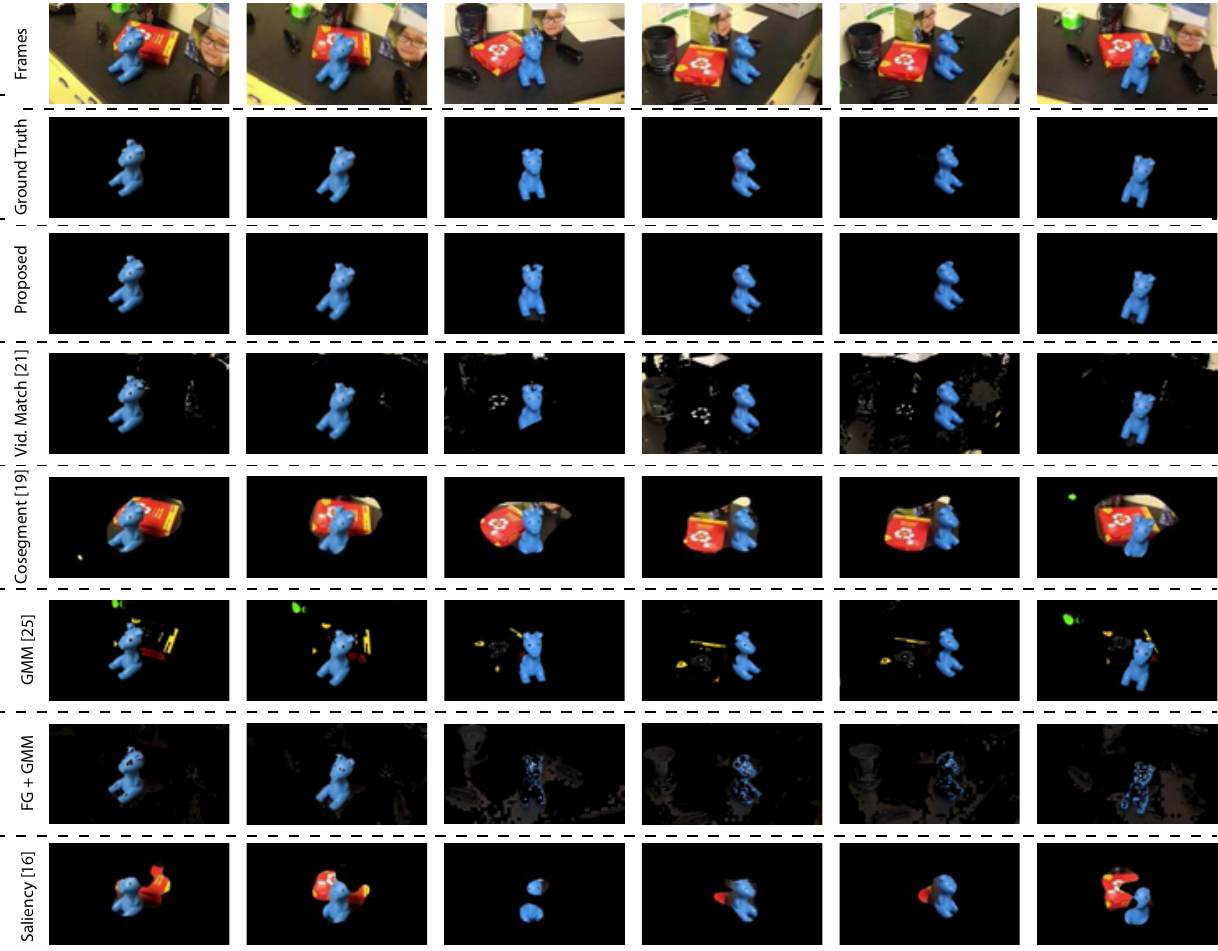}
\label{figur:4}\includegraphics[width=88mm]{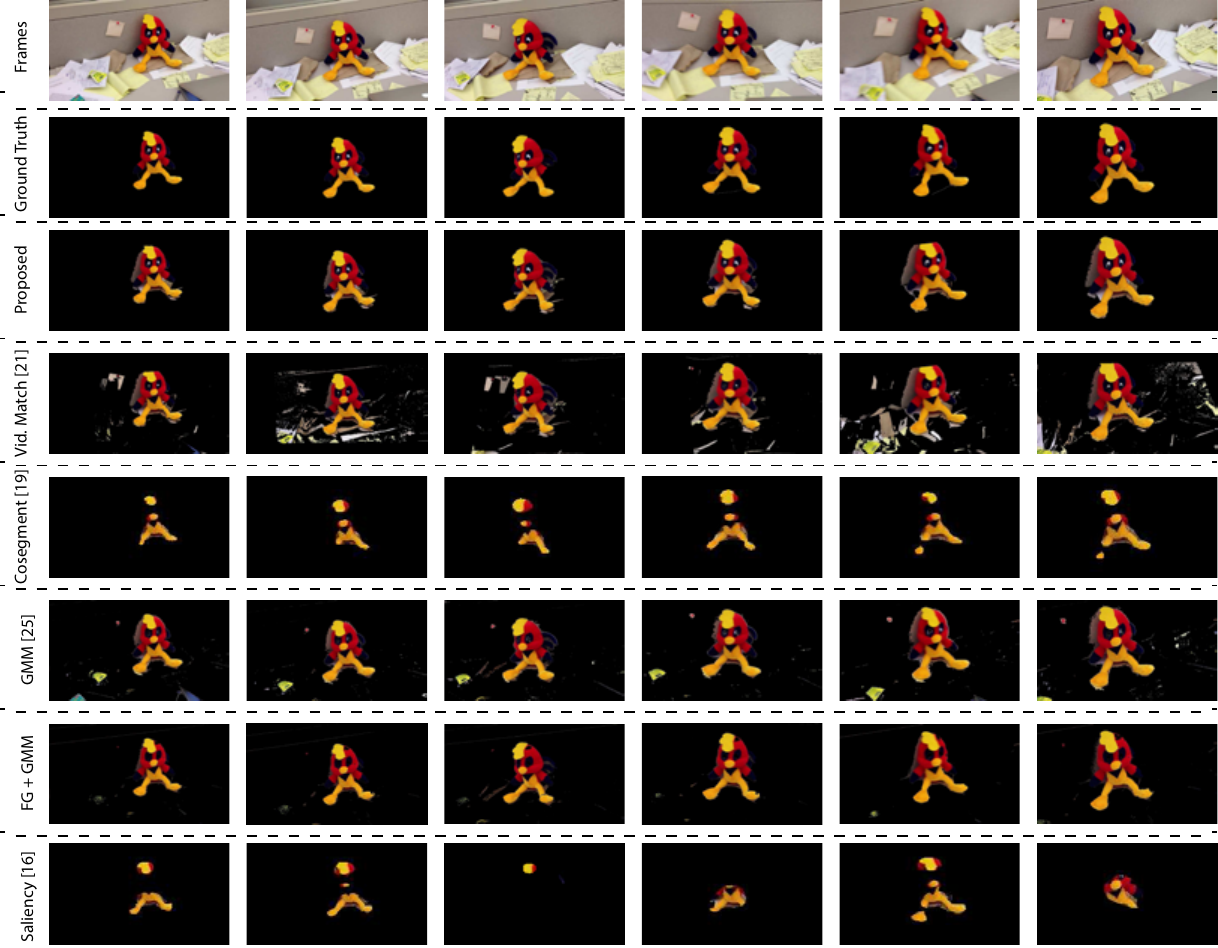}
 --------------------------------------------------------------------------------------------------------------------------------------------
  \\
\label{figur:5}\includegraphics[width=88mm]{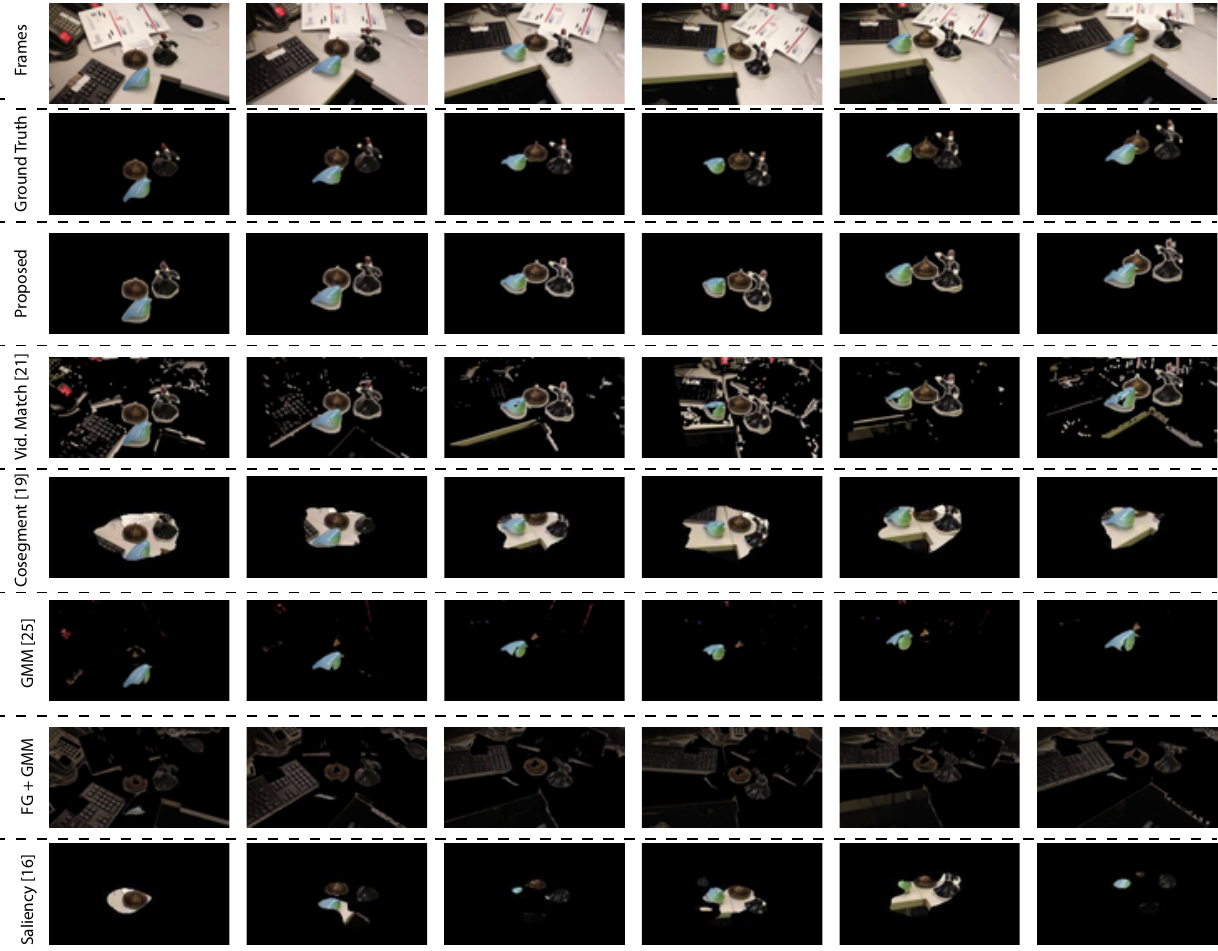}
\label{figur:6}\includegraphics[width=88mm]{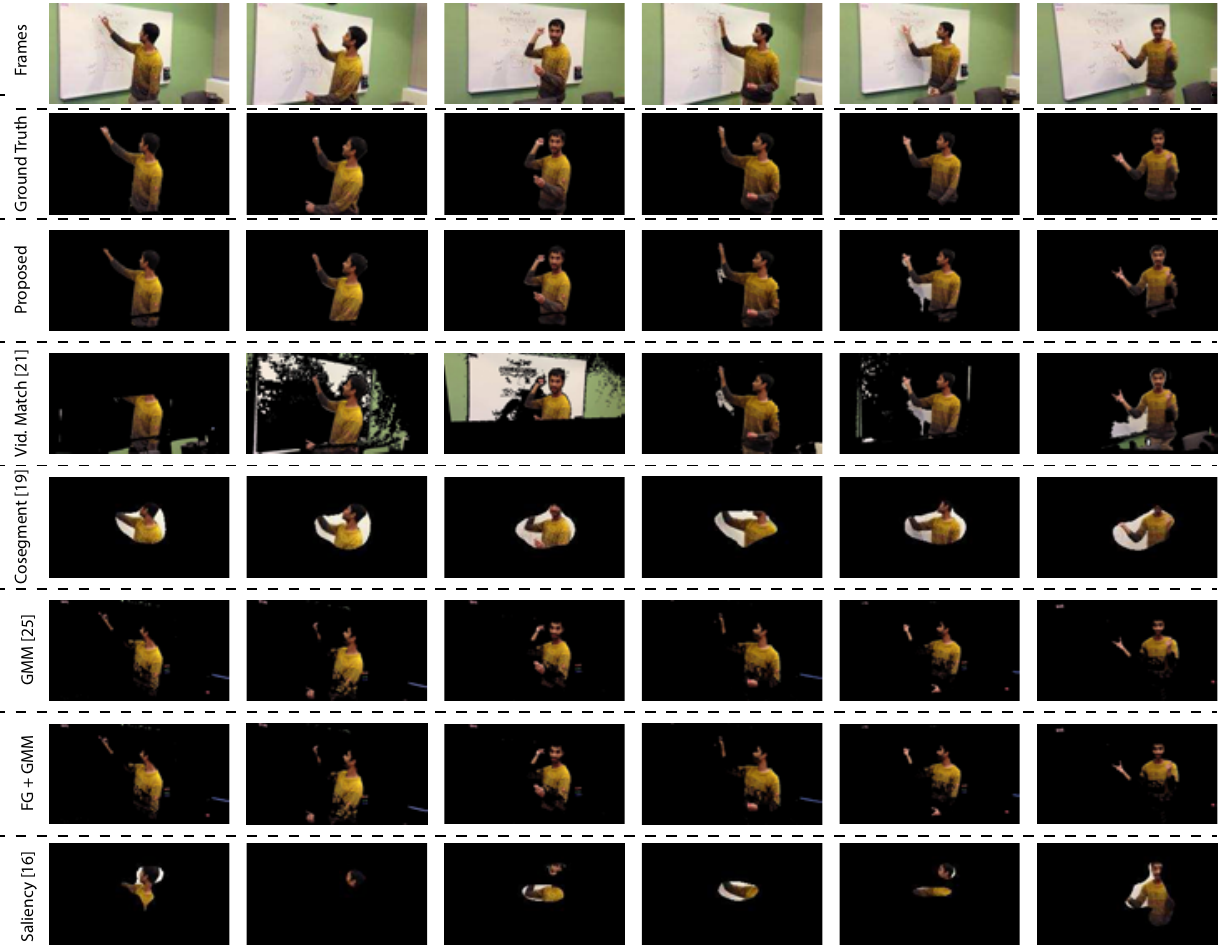}
  \vspace{0.2cm}\caption{Comparative visualizations of subset of frames for some of the test videos are shown. As can be seen, the quality of the output for the proposed method is very similar to that of the manually extracted ground-truth. Also note that the quality of the proposed method is far superior than any of the other comparative algorithm.}
\label{fig:figur}
\end{figure*}

\noindent \textbf{Lemma 3.5} Given a good match between $f$ and $b_i$, the expected number of good matches for $f$ in the background sequence range $[b_{i-\gamma},b_{i+\gamma}]$ is at least $2(\gamma+1)(1-\frac{\delta\gamma}{2\Psi}) - 1$.\\

\noindent \textbf{Proof}: From Lemma $3.4$, we know that if $b_i$ is a good match for $f$, then the probability that $b_j$ is a good match for $f$ is at least $\frac{\Psi-\delta|j-i|}{\Psi}$. Now by using linearity of expectation for all $j$ in range $[i-\gamma,i+\gamma]$, we get that the number of expected matches for $f$ in this range for $b$ is at least\vspace{0.3cm}

\noindent $\sum_{j=i-\gamma}^{i+\gamma}{\frac{\Psi-\delta|j-i|}{\Psi}}$\\
$= 1 + 2\cdot\sum_{j=i+1}^{i+\gamma}{\frac{\Psi-\delta(j-i)}{\Psi}}$\\
$= 1 + 2(\gamma - \frac{\delta}{\Psi}(1+2+\ldots+\gamma))$\\
$= 1 + 2(\gamma - \frac{\delta}{\Psi}\cdot\frac{\gamma(\gamma+1)}{2})$\\
$= (2\gamma+1) - \frac{\delta\gamma(\gamma+1)}{\Psi}$ \\
$= 2(\gamma+1)(1-\frac{\delta\gamma}{2\Psi}) - 1$.\\

\noindent \textbf{Theorem 3.6} Given a good match, with a window size $\gamma = \frac{\Psi}{\delta}$, the expected number of good matches in the entire window of size $(2\gamma+1)$ is at least $\gamma$.\\

\noindent \textbf{Proof}: From Lemma $3.5$, given a good match $b_i$, the expected number of good matches between $b_{i-\gamma}$ and $b_{i+\gamma}$ is at least $2(\gamma+1)(1-\frac{\delta\gamma}{2\Psi}) - 1$. Using $\gamma = \frac{\Psi}{\delta}$ the theorem follows.\\


\noindent \textbf{Theorem 4.1}  Given $r$ good matches, and a chosen voting threshold $t$ in {\sc Voting-Scheme}, the probability that any pixel is classified correctly is at least $1-e^{-O(r)}$.\\

\noindent \textbf{Proof}: Given independent identically distributed random variables $X_i$ that take on values $0$ or $1$, define $X=\sum_{i} X_i$ and let $\mu$ be the expectation of $X$. Then, Chernoff bound states that  $P(|X - \mu|\geq \alpha\cdot\mu) \leq e^{(-\frac{\mu\alpha^2}{2})}$. If a pixel is originally an object pixel, then in the observed matches, it has probability $\geq p_1$. Therefore, if each $X_i$ is defined as the random variable for whether in a good match $i$, the pixel gets marked as an object, then we have $\mu = E(X) \geq p_1r$. So the probability that it gets marked as an object in fewer than $tr$ pixels is $P(|X-p_1r|\geq \frac{|t-p_1|}{p_1}(p_1r))\leq e^{-p_1r(\frac{t-p_1}{p_1})^2/2}$. This is $\leq e^{-(\frac{(t-p_1)^2}{2p_1})\cdot r}$. Now suppose that a pixel was originally a non-object pixel, then in the observed matches, it is classified as a non-pixel with probability $\geq p_2$. Therefore if each $X_i$ is defined as the random variable for whether in a good match $i$, the pixel is incorrectly marked as an object, then we have $\mu = E(X) \leq (1-p_2)r = (p_1-\beta)r$ for some constant $\beta = (p_1+p_2-1) > 0$. Using similar argument as the previous paragraph, we get the probability that the pixel gets marked incorrectly as an object is at most $e^{-(\frac{(t-(p_1-\beta))^2}{2(p_1-\beta)})\cdot r}$. Setting $t$ to any value between $(p_1-\beta)$ and $p_1$, for example $t = p_1 - \frac{\beta}{2}$, we have both quantities $(t-p_1)^2$ and $(t-(p_1-\beta))^2$ are constants {\em strictly} bigger than $0$. It follows that the probability of error in both cases is at most $e^{-(\frac{\beta^2}{8p_1})\cdot r}$ which is $e^{-O(r)}$. The theorem follows.

\section{Result Visualizations}
\noindent For the sake of visual comparison, we show the results of considered approaches on a subset of frames for six of the considered test videos in Figure~\ref{fig:figur}. The results for the remaining test videos show similar visual trends.

\bibliographystyle{abbrv}
\bibliography{egbib}
\end{document}